\documentclass{article}
\usepackage{ijcai20}

\usepackage{times}
\usepackage{soul}
\usepackage{url}
\usepackage[hidelinks]{hyperref}
\usepackage[utf8]{inputenc}
\usepackage[small]{caption}
\usepackage{graphicx}
\usepackage{amsmath}
\usepackage{amsthm}
\usepackage{booktabs}
\usepackage{algorithm}
\usepackage{algorithmic}
\author{
    Nicholas Hoernle$^1$\footnote{Contact Author} \and Kobi Gal$^{1,2}$ \and Barbara Grosz$^3$ \and Leilah Lyons$^4$ \and Ada Ren$^5$ \And Andee Rubin$^5$
    \affiliations
    $^1$University of Edinburgh\\
    $^2$Ben-Gurion University\\
    $^3$Harvard University\\
    $^4$NYSCI\\
    $^5$TERC
    \emails
    n.s.hoernle@sms.ed.ac.uk,
    \{gal,grosz\}@eecs.harvard.edu,
    llyons@nysci.org, 
    \{ada\_ren, andee\_rubin\}@terc.com
}

\title{Interpretable Models for Understanding Immersive Simulations} 

\newcount\Comments  
\usepackage{xcolor}
\newcommand{\kibitz}[2]{\ifnum\Comments=1{\textcolor{#1}{#2}}\fi}

\newcommand{\citename}[1]{\citeauthor{#1}~\shortcite{#1}}

\begin{document}

\maketitle

\begin{abstract} 
This paper describes methods for comparative evaluation of the interpretability of models of high dimensional time series data inferred by unsupervised machine learning algorithms. The time series data used in this investigation were logs from an immersive simulation like those commonly used in education and healthcare training. The structures learnt by the models provide representations of participants' activities in the simulation which are intended to be meaningful to people's interpretation. To choose the model that induces the best representation, we designed two interpretability tests, each of which evaluates the extent to which a model’s output aligns with people’s expectations or intuitions of what has occurred in the simulation. We compared the performance of the models on these interpretability tests to their performance on statistical information criteria. We show that the models that optimize interpretability quality differ from those that optimize (statistical) information theoretic criteria. Furthermore, we found that a model using a fully Bayesian approach performed well on both the statistical and human-interpretability measures. The Bayesian approach is a good candidate for fully automated model selection, i.e., when direct empirical investigations of interpretability are costly or infeasible.
\end{abstract}

\section{Introduction}
\label{sec:introduction}
This paper investigates methods for evaluating the interpretability of models of time series data arising  from people's interactions in immersive simulations  
such as those  used for teaching in healthcare, disaster response and science education~\cite{alinier2014immersive,amir2013plan}.
In such simulations, people's interactions  engender a rich array of emergent outcomes and yield diverse opportunities for learning~\cite{smordal2012hybrid}. In the immersive simulation used in this study, Connected Worlds\footnote{Installed at the New York Hall of Science (NYSCI): \url{https://nysci.org/home/exhibits/connected-worlds/}} (CW), students interact with an ecological simulation to learn about the causal effects of their actions on environments over time~\cite{mallavarapu2019connect}.

Rich causal relationships, simultaneous participation from students and the changing dynamics of immersive simulations can make it difficult for people to determine how their interactions with the simulation caused the changes they observe in the simulated world.
Machine learning methods can be used to summarize the effects of participants' actions over various time periods. 
For such methods to be effective, though, they must meet the challenge of identifying a model that is both ``true to the data'' and understandable to the target audience interested in uncovering the causal relationships.

This paper defines and solves the \emph{interpretability problem for immersive simulation settings}: determining that machine learning model, from a set of candidates, that people understand best~\cite{doshi2017roadmap,caruana2015intelligible}. 
It compares the selection of a model according to a criterion that optimizes for maximum statistical information with one that optimizes for interpretability.
The ability to identify the model that is best (or among the top choices) for interpretability is essential to a system's capability to explain its conclusions~\cite{rosenfeld2019explainability}.

Our approach to addressing the interpretability problem comprises the following:
(1) select a set of machine learning models for segmenting time series data; in the domain we investigated, the segmentation is of students' interactions with CW into coherent periods of time; 
(2) design tests for computing the interpretability score of a model for a given input; 
(3) empirically evaluate the models with respect to their interpretability score in a user study. 

To infer the boundaries of stable periods in the data of CW dynamics, we use a family of hidden Markov models (HMMs).
These HMMs are augmented with an additional ``sticky'' hyperparameter which biases the transition dynamics of the latent state-space~\cite{fox2008hdp}.  
The input to each HMM is a multidimensional time series representing the response of the CW system to actions performed by students in the simulation. 
The output of the HMM is a segmentation of the time series into a set of periods, which are contiguous lengths of time during which the system dynamics form a stable linear process. 

We implemented two tests of interpretability for CW models: the Forward Simulation and Binary Forced Choice \cite{doshi2017roadmap}. 
These tests each determine the extent to which the learnt representations are interpretable to people, albeit in different ways. 
They both use a visualization of the inferred periods that shows experimental subjects snapshots of the CW system's state from the selected periods that the HMM inferred.
 
The results showed that the interpretability of the different models varied according to the value(s) of HMM parameters. 
In particular, the HMM that optimized statistical information criteria did not optimize interpretability quality.
In addition, a fully Bayesian approach, which does not require hyperparameter tuning, offered a good balance between interpretability and performance on the theoretical statistical tests.
We argue that the Bayesian approach could be suitable for situations in which it is not possible to engage people in determining interpretability or doing so would be unethical or impractical. 

This paper makes three contributions. 
First, it provides an end-to-end paradigm for the design and evaluation of the interpretability of models for unsupervised learning in time series domains.
Second, it defines new interpretability tests for unsupervised time-series settings, and applies them to real-world data. 
Third, in identifying the Bayesian solution, it provides an attractive alternative to model selection when human subject experimentation is not possible.
Finally, we note that the results of this investigation have been deployed in a classroom study for the purpose of assisting teachers in explaining systems thinking to students who participated in the CW simulation study.

\section{Related Work}

\citename{doshi2017roadmap} suggested three tests to evaluate how interpretable a model's representations are to people. 
Forward Simulation: requires a human evaluator to predict the output of a model for a given input. 
Binary Forced Choice: requires an evaluator to choose one of two plausible model explanations for a data instance.
Counterfactual Simulation: requires an evaluator to identify what must be changed in an explanation to correct it for a given data instance.

In follow-up work~\citename{lage2018human} propose a model selection process that considers both a model's accuracy and its degree of interpretability, according to one of the above tests. 
They provide a framework for iteratively optimizing the interpretability of a model with a human-in-the-loop optimization procedure. 
Their work applied this framework to tests in the lab in which human judgment was used to optimize supervised learning models.
Other works that studied interpretability tests for supervised learning settings include~\citename{wu2018beyond,ribeiro2016should,choi2016retain,lipton2016mythos}.
We extend this literature on interpretability by adapting the model selection process to an unsupervised learning setting, that of segmenting a multi-dimensional time series into periods. 
Moreover, we implement examples of the Forward Simulation and Binary Forced Choice tests suggested by~\citename{doshi2017roadmap} and apply them to a high dimensional time series setting.

Our work was inspired by \citename{chang2009reading} who were the first to show that optimizing machine learning models in unsupervised settings using predictive log-likelihood may not induce models that are interpretable to people.
They focused on the use of topic models for finding meaningful structure in documents and they compared the models that are selected to optimize \textit{perplexity} (analogous to held-out log-likelihood) to the models that were selected by the human interpretability tests that they designed. 
\citename{chang2009reading} operationalized two Forward Simulation tests for evaluating the interpretability of a topic model: word intrusion, in which the evaluator is required to identify which of several words does not belong together in one topic represented by the other words; and  topic intrusion, in which the evaluator is required to 
identify which of several topics is not associated with a given document.
We extend this work to a multi-dimensional time series domain and we introduce a Binary Forced Choice test to complement the ``intrusion'' Forward Simulation test.


\section{The Connected Worlds Domain}
\label{sec:cw_desc}
Connected Worlds (CW), a multi-person ecology simulation (installed at NYSCI), aims to teach students about complex systems and systems thinking. Its immersive environment comprises four biomes (Desert, Grasslands, Jungle \& Wetlands) connected by a central water flow fed by a waterfall. 
Students plant trees which flourish or die, animals arrive or depart, and rain clouds form and rain feeds the waterfall. 

\begin{figure}[t]
\centering
\includegraphics[width=0.39\textwidth]{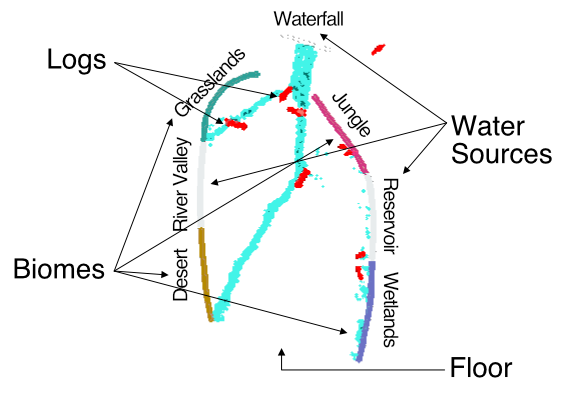}
\caption{CW session-view. Biomes are labelled on the perimeter and logs appear as thick red lines. Water (blue stream in the middle of the image) enters via the waterfall and in this image it mainly flows toward the Grasslands and the Desert.}
\label{fig:connected_worlds_graphic}
\end{figure}

Students control the direction of water flows in the simulation by moving foam logs to direct water among the biomes. 
Water enters the simulation through rainfall events, which are not under student control.  
Figure~\ref{fig:connected_worlds_graphic} gives a snapshot of the system state, we refer to this snapshot as the session-view. 
This session-view is a system-generated representation of the water flows and it directly reflects the logged water flows and levels in the simulation.
The output of a CW session is a time series recording the levels of water in the different biomes for 8 minutes at a 1Hz frequency.
The ability to model the effects of student actions on the environment were limited by two factors: The time series was the only source of information about students' interactions, and it was not possible to access the CW simulation except at NYSCI.

The CW simulation is complex on several dimensions as a large number of students simultaneously execute actions that change the state of the simulated environment. 
Each participant has a different view of what transpired, depending on the actions s/he took and the environment changes that resulted. 
Students' activities are recorded as a movie (see Figure~\ref{fig:connected_worlds_graphic}) that can be shown to students and teachers. 
This movie can inform discussions about the causal effects of the students' actions on simulation outcomes, but it obscures temporal dependencies in their interactions.
This limitation motivated the use of ML algorithms to better support students' understanding of the effects of their actions on the simulation's progression.


\section{Interpretability Tests for CW}
\label{sec:interpretability_tests}

\begin{figure}
\centering
\includegraphics[width=0.42\textwidth]{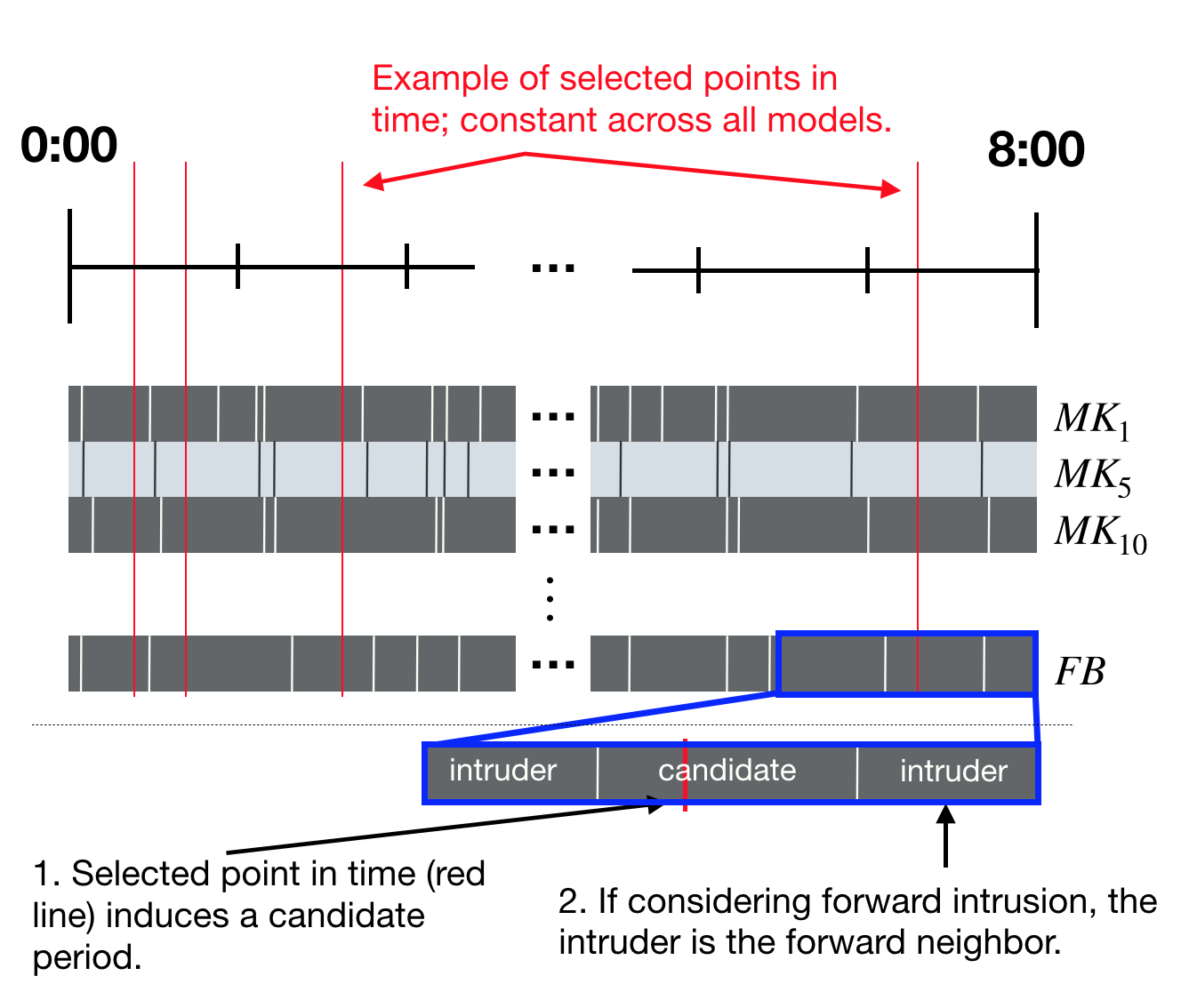}
\caption{
The time series is represented as a horizontal line from minute 0 to 8; red vertical lines denote sampled time points in the time series; each model is shown as a grey rectangle; models segment time series into periods delimited by white vertical lines. The forward or backward neighbour of the candidate period is selected as an intruder.}
\label{fig:selection_representation}
\end{figure}

\begin{figure*}[t]
\centering
\includegraphics[width=.85\textwidth]{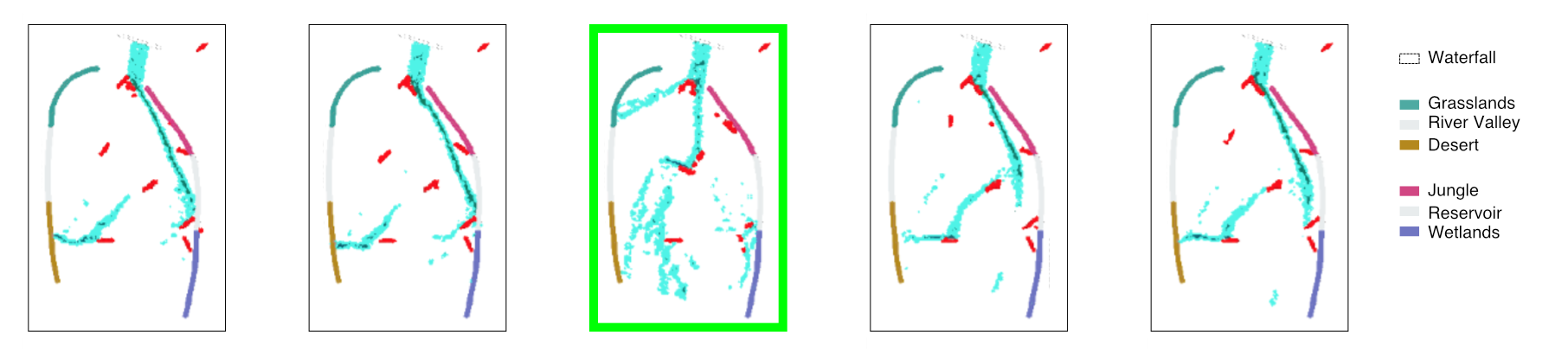}
\caption{Screenshot of the Forward Simulation test interface. Here $4$ of the images show water flowing towards the Desert. An intruder image, the highlighted one, comes from a different period and shows water flowing to both the Desert and the Grasslands.}
\label{fig:test1_screenshot}
\end{figure*}

Let $D$ be a time series that records the levels of water in the different biomes. 
Let $M$ be a model that takes as input a time series $D$ and outputs a segmentation of $D$ into periods. 
Each period aims to provide a coherent description of the water flow for a length of a time. 

Importantly, a single period is insufficient for modeling the effects of students' interactions with CW, because students' sustained actions have complex effects on the system dynamics over time. 
For example, when students choose to direct water to the Desert and Plains and plant trees in the Desert, the system dynamics are entirely different from the case when water is directed towards the Jungle and the Desert, and the Plains are left to dry. 
We must therefore allow for multiple periods. 
Each period describes a length of time where water flowed to a sufficiently stable target. 
From the above example, one period can describe water that mainly flows to the Plains and to the Desert; students then move logs to re-route water flow to the Jungle, thus starting a new period. 

We use an interpretability score $IS$ to measure the interpretability of a model $M$ applied to $D$.  
The interpretability score is computed via an average across test instances, $T(M,D,i)$, which each take as input a model $M$, a time series $D$ and a selected point in time $i$ from the time series.
Each test instance returns True if an evaluator successfully completes a required objective.

We adapted the Forward Simulation and Binary Forced Choice tests~\cite{doshi2017roadmap} to the CW domain using the notion of \textit{candidate} and \textit{intrusion} periods.  
We say that period $p$ is \textit{active} for model $M$ at time $i$ if $M$ infers the period $p$ to describe a contiguous length of time in the time series, and $p$ includes the time $i$.
Figure~\ref{fig:selection_representation} shows how the tests select candidate and intrusion periods. 
First, a time point (red vertical line) is used to select a candidate period where the candidate period is the active period from model $M$ at $i$ (the active period for a model intersects with the red line). 
Then, the intrusion period is selected as a direct neighbor to a candidate. 
Each test is operationalized via a \emph{visualization} which presents any period as a set of images extracted from the session-view. 

Figure~\ref{fig:test1_screenshot} shows an example of the Forward Simulation test on a real data instance.
As shown by this figure, the test sampled $4$ session-view images from the candidate period of model $M$ at time $i$, and a single session-view image sampled from the intrusion period. 
The images were presented in a random order.
In Figure~\ref{fig:test1_screenshot}, the image that is outlined in green is the intrusion image that corresponds to the intrusion period.  
A test evaluator was required to identify which image was the intrusion image. 

\begin{figure*}[t]
\centering
\includegraphics[width=0.62\textwidth]{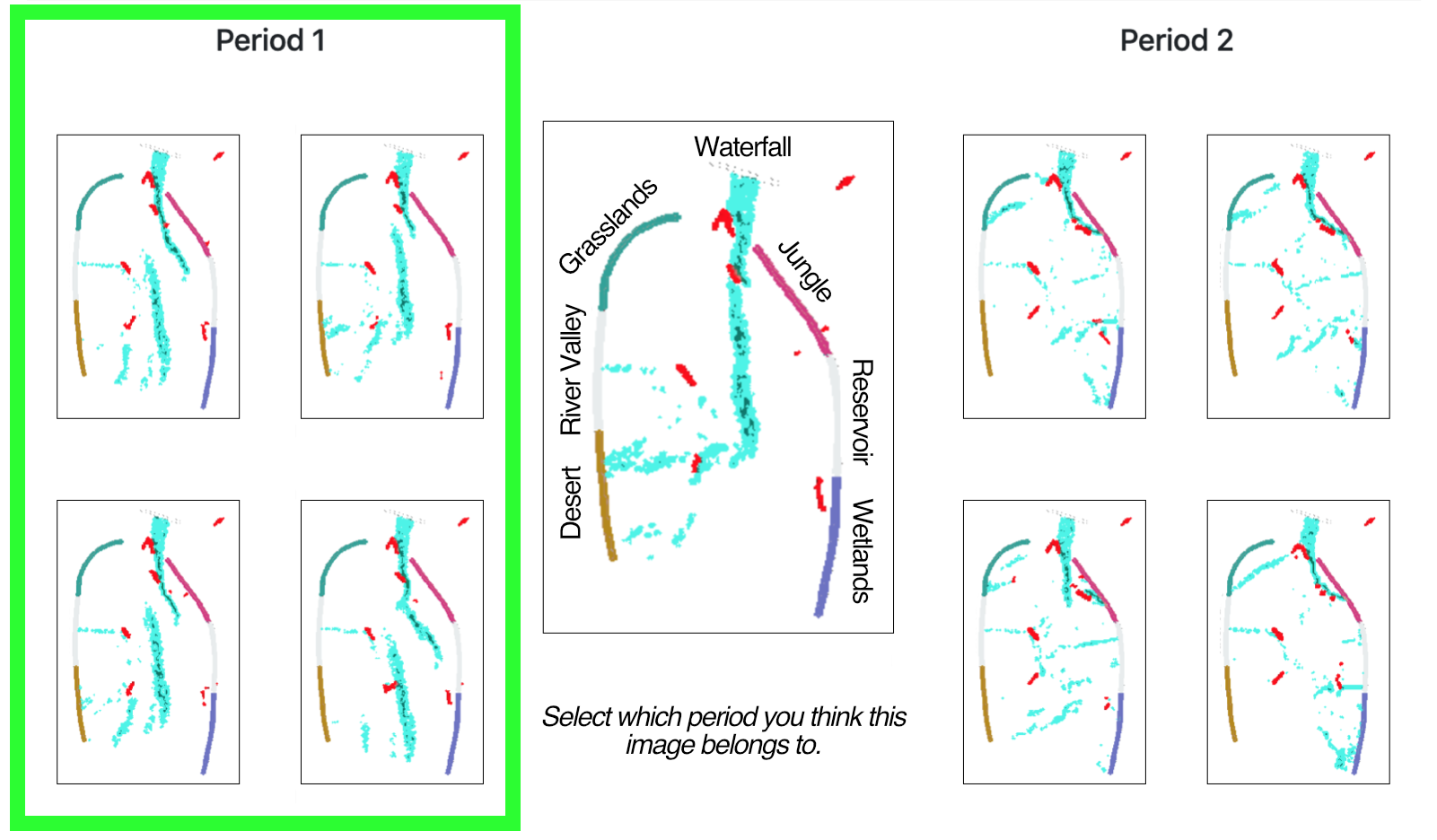}
\caption{Screenshot of the Binary Forced Choice user interface. An unknown center image needs to be associated with either ``Period 1'' or ``Period 2''. In this case, streams of water flowing to both the Grasslands and to the Jungle capture the dynamics in Period 2. Period 1 has a small amount of water reaching the Desert which is consistent with the unknown image.}
\label{fig:test2_screenshot}
\end{figure*}

Figure~\ref{fig:test2_screenshot} presents an example of the Binary Forced Choice test. 
The test displays an unknown session-view image from a candidate period (center of screen) and additional images from two competing periods that contain this image (``Period 1'' or ``Period 2''). 
Each of the two competing periods is visualized as four images sampled from the candidate or the intruder period. 
The unknown image is sampled in time close to the boundary of when the candidate period transitions into the intruder period.
In Figure~\ref{fig:test2_screenshot}, Period 1, highlighted in green, is the period that correctly explains the unknown image (i.e., the images in ``Period 1'' and the ``unknown image'' are all sampled from the candidate period).
A test evaluator is required to choose between the two possible periods.

Hypothetically, the intruder period can be chosen arbitrarily, as in~\citename{chang2009reading}.
However, intrusion periods that are further away in time from the candidate period would be easier to detect due to the non-stationary evolution of the system. 
We made a design decision to chose the period that is immediately adjacent to the candidate period, either forward or backward in time. 
This makes it harder to distinguish between candidate and intrusion period, but provides a rigorous test for the specific choice of boundary between the two periods. 

Given data set $D$ and model $M$, the interpretability score $IS$ of a model is equal to the average success of the test instances for model $M$ over multiple points $\{i\}$ in a time series $D$.
The set of time points $\{i\}$ were uniformly sampled from the time series with the additional constraint that each minute of interaction had at least one sample.
For every model we test, we hold constant the selected times $\{i\}$ in the time series (as shown in Figure~\ref{fig:selection_representation}). 
In this way we control for different areas in the time series being more or less difficult to segment into coherent periods.
 
\section{Modeling Students' Activities in CW}
\label{sec:model_for_segmenting_time}

In this section we describe the design of general models for segmenting students' activities into periods of time and thereafter present the specific classes of model that are used in our interpretability tests.

\subsection{Segmenting Time Series Data into Periods}
\citename{hoernle2018modeling} used a HMM to model the system responses to students' activities in CW in which the latent states of the HMM corresponded to periods. 
Transitions between different states equates to the system changing between different periods, while self transitions mean the system persists within the same period. 
The authors did not address the question of how to choose the number of states. To this end, we augment the HMM with a hierarchical Dirichlet process which places this non-parametric prior over the state space, following the approach detailed by \citename{teh2005sharing} and \citename{fox2008hdp}.

The ``Sticky-HMM" approach introduced by \citename{fox2008hdp} includes a hyperparameter, $\kappa$, that biases the model to persist in a state, given that it has already adopted that state. 
Applied to CW, the greater the value for $\kappa$, the more the model will try to persist in any given state. 
The increase in the length of periods corresponds to a decrease in the number of latent states. 
The opposite is true for lower values of $\kappa$ where there is a lower bias to persist within a given state and consequently there are more periods that are inferred.
For a detailed description of the model, including the  Gibbs sampling inference scheme that is used to infer the model parameters, refer to \citename{fox2008hdp} and \citename{fox2009bayesian}. 

\subsection{Model Classes}
We introduce three classes of model that segment time into periods that can be used to explain the water flows:

\begin{enumerate}
    \item  {${MK_X}$: sticky HMM with fixed $\kappa$.} We use the basic structure of the sticky HMM described by \citename{fox2008hdp} with set values for $\kappa$ to produce $10$ unique models, spanning a wide range of possible settings\footnote{$\kappa \in \{1, 5, 10, 50, 100, 150, 200, 300, 500, 700\}$.}.  
    \item  {${FB}$: fully Bayesian sticky HMM with Gamma prior on $\kappa$.} This  approach places a weakly informative, conjugate Gamma prior on the hyperparameter that expresses high uncertainty over the $\kappa$ values\footnote{The \textit{(shape,rate)} parameters were chosen to be $(1,\frac{1}{4})$; empirical results were invariant to a range of these values.}.
    \item  {${Rand}$: Random baseline.} The random baseline generates periods of random length drawn from a Poisson distribution with mean set to be the mean of all other periods induced by the parametric models. The random periods are defined to include the selected time points ($\{i\}$ from Section~\ref{sec:interpretability_tests}).
\end{enumerate}

\noindent We refer to ${FB}$ as the fully Bayesian model to indicate the fact that the none of the parameters of interest are specified and consequently posterior inference is over all of the parameters in the model (including $\kappa$). This is in contrast to the ${MK_X}$ models where we explicitly set the value for the sticky parameter $\kappa$.

For models in class $1$ and $2$, we use the Gibbs sampler, described by \citename{fox2008hdp}, to perform inference over the parameters in the model, this includes inference over the state sequence and thus the period segmentation of the model. The observation distribution was chosen to be a mixture of two multivariate Gaussians with conjugate Normal-inverse-Wishart priors. This mixture model addresses the noise in the CW water flow, such as ``splashes'', which prior work has identified as a challenge in this domain~\cite{hoernle2018modeling}.

\section{Model Selection for Interpretability}
\label{sec:model_selection}
The goal of model selection is to optimize a metric such that a specific parameter setting can be chosen as the best model for use during inference. 
We compare how the models from section~\ref{sec:model_for_segmenting_time} perform on both statistical tests and on the human interpretability tests outlined in section~\ref{sec:interpretability_tests}. 

\subsection{Selection using Statistical Information}
\label{sec:baseline}
When  human interpretability testing is infeasible,  one could choose to optimize some proxy to interpretability~\cite{doshi2017roadmap,lage2018human}. 
For example, \citename{chang2009reading} compared the proxy of held-out log-likelihood to the human interpretability score that was a result from two tests that were run on Amazon Mechanical Turk (Mturk).

Ideally, the model parameters would be optimized on held-out data using predictive log-likelihood as the objective~\cite{chang2009reading}. 
However, the difficulty of collecting controlled sessions of student interaction in CW meant we had few data instances available (see limitation discussion in the next section). 
To address this challenge we use statistical information criteria as a theoretical approximation to the predictive accuracy of a model~\cite{gelman2013bayesian}.  

\begin{figure}[t]
\centering
\includegraphics[width=0.42\textwidth]{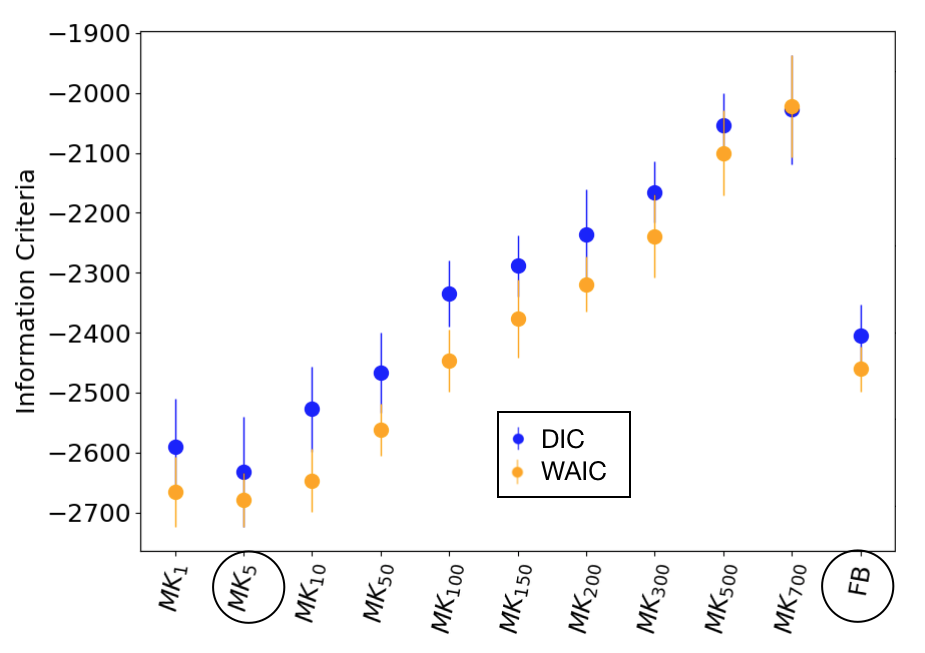}
\caption{DIC and WAIC as a function of the model (lower  is better). The $MK_5$ model is optimal, the $FB$ approach is in 5th place.}
\label{fig:dic_waic_scores}
\end{figure}

Figure~\ref{fig:dic_waic_scores} shows the two information criteria (the Deviance Information Criteria, DIC, and the Watanabe-Akaike Information Criteria, WAIC~\cite{gelman2013bayesian}) plotted as a function of the model (the random model has no notion of information criteria and so was not compared here). The data set comprised of both of the log files of students' interactions (8 minutes each). The optimal model for both DIC and WAIC is the $MK_{5}$ model but we note that $MK_{1}$, $MK_{5}$ and $MK_{10}$ all perform close to this optimal setting. Notice that the fully Bayesian model (FB) is not optimal but it is in the top $5$ models for both criteria. 
 
\subsection{Selection using Interpretability Test} 
\label{sec:interpretability_results}
This section describes the choice of model according to interpretability quality, as measured by the interpretability tests. 
The set of models used in this study includes the 12 CW models described in Section~\ref{sec:model_for_segmenting_time}. 
IRB was obtained for the study.

We recruited participants from two cohorts: undergraduate engineering students in a large public university and Mturk workers (with a total of $240$ people who participated in the experimentation).
For a given time series $D$ in CW, we randomly sampled a set of 12 time points, which remained constant across all model conditions.    
Each time point was used to generate a candidate and two intrusion periods (both forward and backward in time, see Figure~\ref{fig:selection_representation}), making for $2 \times 12 \times 12=288$ tests per time series. 
We divided participants into two cohorts, one for Forward Simulation, and one for Binary Forced Choice tests. 
Both cohorts varied the models used to generate their respective tests. 
Each participant performed $20$ tests, with no more than $2$ tests generated from any given model, to ensure a representative range of models.  
After making their choice, participants received brief visual feedback on whether or not their selection was in agreement with the model's choice.

All participants received a detailed tutorial about CW and the study, as well as a pre-study comprehension quiz\footnote{Tutorial pdf slides are available at \url{https://www.dropbox.com/s/pu2nxk2k0g81ql6/forijcai.pdf}}. Mturk workers were paid a base rate of $\$0.25$ for participating and a bonus structure of $\$0.1$ for each correct response.
 
\begin{figure}[t]
\includegraphics[width=.45\textwidth]{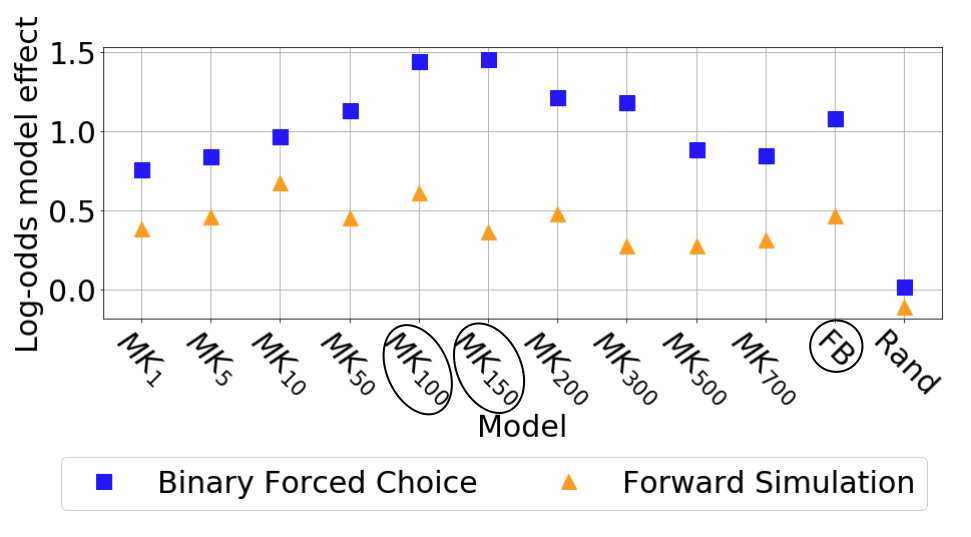}
\caption{Effect of each model on the log-odds of a test evaluator selecting the correct response (controlling for the test evaluator, the experiment trial, log file and ordering effects).}
\label{fig:essil_test_results}
\end{figure}

We first describe results in terms of accuracy (the percent of correctly labelled test instances). 
The top performing model was $MK_{200}$ with an accuracy of $83\%$ on the Forward Simulation  test and $MK_{100}$  with an accuracy of $82\%$ on the Binary Forced Choice test. 
The random baseline model performed consistently poorly with an average accuracy of $53\%$ on both tests. 
The fully Bayesian model achieved an accuracy of $72\%$ and $70\%$ respectively on the two tests. 

To control for ordering effects, chosen time periods, data instance used, and effects of individual participants, we applied an L2 regularized logistic regression for predicting  the user specific success on the experiment trial, shown in  Figure~\ref{fig:essil_test_results}. The y-axis presents the improvement in log-odds that a model has on the expected response accuracy (higher is better). 
As shown by this figure, the Forward Simulation shows a high variance with no clear maximum. In contrast, the Binary Forced Choice test has a clear maximum in the region of $MK_{100}$ and $MK_{150}$.

From Figures \ref{fig:dic_waic_scores} and \ref{fig:essil_test_results} we can infer the following four conclusions.
First, all of the models ($MK_{1},\ldots,MK_{700}, FB$) outperform the random baseline: participants are more likely to select the correct response from any of these models. This result suggests that periods of stable dynamics exist in the data and that it is possible to construct models, which describe these dynamics, that are interpretable to people. 

Second, the Binary Forced Choice test is a preferable measure for interpretability to the Forward Simulation test. Figure~\ref{fig:essil_test_results} shows that the Binary Forced Choice test exhibits a clear peak (around $MK_{100}$ and $MK_{150}$) where interpretability of the model is maximized. 
These models also maximized the  raw accuracy on the Binary Forced Choice test.

On the other hand, the Forward Simulation test has a greater variance across models and across data instances. 
Two possible causes for this higher variance are: (1) there is more room for error in the Forward Simulation test (5 choices vs. 2 choices in Binary Forced Choice); (2) sampling a single image to represent a period (as in Forward Simulation) presents less information to the user than sampling 4 images (as in Binary Forced Choice). 

Third, the best $\kappa$ settings vary for different tests and information criteria. 
Model interpretability grows steadily as the value of $\kappa$ increases, with $MK_{100}$ and $MK_{150}$ being the optimal models, and then proceeds to decrease steadily.   
These models are not consistent with the model $MK_5$ that optimized the information criteria.
Note that higher $\kappa$ values are ``sticky'' - they  bias the model towards longer periods, which condense too many activities to make sense to people. 
On the other end of the spectrum, lower $\kappa$ values allow for more (shorter) periods that may  capture  noise in the system.   The $\kappa$ value for  models $MK_{100}$ and $MK_{150}$ represent a ``sweet spot" in between these two extremes.

Finally, the fully Bayesian model $(FB)$ performs consistently well on both information criteria and interpretability tests. It is interesting to note that while this model does not find the optimal setting (from neither the statistical information criteria nor from the human interpretability task) it does perform well across all tests, tasks and instances, and is fully automated (no human evaluation is required in order to choose an optimal parameter setting).

We conclude this section with mentioning the limitation that the user study was based on a small number $(n=2)$  of instances. This was due to the difficulty in obtaining controlled sessions of student behavior in CW. Despite this issue, the differences between the models in Figure~\ref{fig:essil_test_results} are statistically significant, having being evaluated across 12 different time points for each instance and with hundreds of evaluators.

\section{Conclusion \& Future Work}
With the growing prevalence of immersive simulations the need arises for AI systems which help people gain insight into the ways participants' activities affect the simulation outcomes. 
We have studied an environmental simulation intended to teach students  about the causal effects of their actions. 
Our results show that algorithms can segment time series log data into periods that are meaningful for people. 
Selecting hyperparameters in these models is a challenge, especially when trying to optimize the representations they produce for their interpretability.
We have described ways to select these hyperparameters using two tests that are grounded in the literature. 
We showed that the fully Bayesian method is a promising technique for implementing a model when people cannot directly assess and evaluate the models. 
Our results are important for any unsupervised machine learning task for which interpretability is an important criterion, because in such cases the model selection problem will be encountered. 
The work forms part of a broader project where the goal is to generate relevant summaries of the CW dynamics such that teachers can effectively engage their students in discussions about their own experiences with the simulation. 

In future work we plan to explore alternative ways to measure interpretability quality in time series domains, including the design of a counterfactual simulation test~\cite{doshi2017roadmap},  and the application of our approach to additional domains. 

\section*{Acknowledgments}

Nicholas Hoernle is supported by a commonwealth scholarship. This paper is based on work supported by the National Science Foundation under grant numbers \#1623091, \#1623124 and \#1623094.
 Thank you to  Michael Rovatsos and  Alex Lascarides  for their advice on this project, to the reviewers for their helpful comments and to NYSCI for assisting with the data collection.

\bibliographystyle{named}
\bibliography{bibliography}
\end{document}